\documentclass{article}

\usepackage[english]{babel}
\usepackage{url}
\usepackage{nicematrix}
\usepackage{multirow}
\usepackage{colortbl}
\usepackage{array}
\usepackage{enumitem}
\usepackage{comment}
\usepackage{algorithm}
\usepackage{algpseudocode}
\usepackage{lineno}
\usepackage{authblk}
\usepackage[normalem]{ulem}
\usepackage[dvipsnames]{xcolor}
\usepackage[margin=1in]{geometry}
\usepackage{amsmath}
\usepackage{cite}
\usepackage{amssymb}
\usepackage{mathtools}
\usepackage{graphicx}
\usepackage[colorlinks=true, allcolors=blue]{hyperref}
\usepackage[capitalize]{cleveref}
\usepackage{fancyhdr}

\newcommand{\sid}[1]{\textcolor{red}{{\em {\bf Sid:} #1}}}

\title{Imaging Hidden Objects with Consumer LiDAR \\ via Motion Induced Sampling}
\author{Siddharth Somasundaram$^1$, Aaron Young$^1$, \\ Akshat Dave$^1$, Adithya Pediredla$^2$, Ramesh Raskar$^1$}
\affil{$^1$Massachusetts Institute of Technology, Cambridge, MA, United States \\  $^2$Dartmouth College, Hanover, NH, United States}

\usepackage{docmute}

\begin{document}

\let\origbibliography\bibliography
\let\origbibliographystyle\bibliographystyle
\renewcommand{\bibliographystyle}[1]{}
\renewcommand{\bibliography}[1]{}

\def\maketitle{%
  \begin{center}
    {\LARGE\bfseries Imaging Hidden Objects with Consumer LiDAR \\ via Motion Induced Sampling\par}
    \vspace{1em}
    {\large Siddharth Somasundaram$^1$, Aaron Young$^1$, \\ Akshat Dave$^1$, Adithya Pediredla$^2$, Ramesh Raskar$^1$\par}
    \vspace{0.5em}
    {\footnotesize $^1$Massachusetts Institute of Technology, Cambridge, MA, United States \\ $^2$Dartmouth College, Hanover, NH, United States\par}
    \vspace{2em}
  \end{center}
}

\date{}
\maketitle
\thispagestyle{fancy}



\vspace{-3em}

\begin{center}
\fbox{%
  \begin{minipage}{0.52\linewidth}
    \centering
    {\large Project Page:}
    \href{https://sidsoma.com/consumer-nlos/}{\large sidsoma.com/consumer-nlos/}
  \end{minipage}%
}
\end{center}


\begin{abstract}
\noindent LiDARs are being increasingly deployed for consumer imaging in handheld, wearable, and robotic applications. These sensors can capture the time-of-flight of light at picosecond resolution, which in principle, enables them to capture information about objects hidden from their field of view. While such non-line-of-sight (NLOS) imaging capabilities have been shown on research-grade LiDARs, they are challenging to achieve on consumer devices due to poor signal quality resulting from low laser power, low spatial resolution, and object and camera motion. Inspired by burst photography and synthetic aperture radar, we propose a multi-frame fusion strategy to overcome these challenges and demonstrate NLOS imaging on consumer LiDAR. We first introduce the motion-induced aperture sampling model to unify the effects of object shape, object motion, and camera motion under a single measurement model. Using this model, we demonstrate several NLOS capabilities on a smartphone-grade LiDAR: (1) 3D reconstruction, (2) single and multi-object tracking, and (3) camera localization using hidden objects. Previously, NLOS imaging capabilities were largely restricted to bulky and expensive research-grade hardware that requires extensive setup and calibration. Our results represent a shift towards \textit{plug-and-play NLOS imaging}, where anyone can image hidden objects with off-the-shelf hardware ($<100\$$) and no additional setup. We believe that democratization of such capabilities will advance consumer applications of NLOS imaging.

\end{abstract}

\section{Introduction}

\begin{figure}
    \centering
    \includegraphics[width=0.9\linewidth]{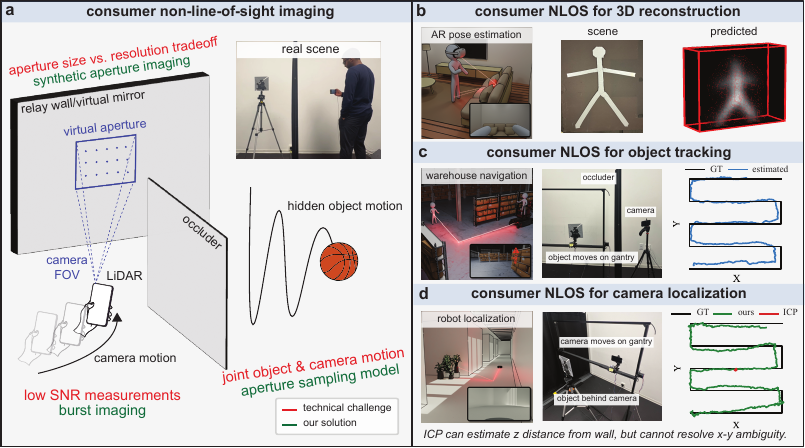}
    \caption{\textbf{Consumer non-line-of-sight imaging}. (a) NLOS imaging is possible by turning nearby diffuse surfaces into virtual mirrors that reveal hidden objects. However, NLOS imaging with consumer LiDARs is challenging due to low SNR, the tradeoff between virtual aperture size and sampling density, and joint object and camera motion. We address these challenges using multi-frame fusion. (b) We demonstrate three consumer NLOS capabilities---3D reconstruction, tracking, and camera localization---on smartphone-grade LiDAR. The insets of the application figures shows the limitation of only relying on line-of-sight signals. 
    }
    
    \label{fig:teaser}
\end{figure}







Today, most consumer imaging systems---handheld, wearable, or robotic---can only see objects within direct line of sight of the camera. But what if these everyday cameras could also see beyond line of sight (\cref{fig:teaser}a-b)? Augmented reality systems would be able to robustly map the user's body pose for seamless integration into the virtual world. Warehouse robots would be able to navigate fast-paced and unstructured environments while avoiding collisions around the corner. Roombas would remain oriented in textureless spaces by using temporarily hidden objects to localize itself. Motivated by such use cases, we revisit the non-line-of-sight (NLOS) imaging problem through the lens of consumer mobile imaging.




Several approaches have been proposed for NLOS imaging \cite{adib2013see, yue2022cornerradar, scheiner2020seeing, maeda2019thermal, lindell2019acoustic, willomitzer2021fast, batarseh2018passive, freund1990looking, katz2014non, smith2018tracking, bouman2017turning, saunders2019computational, tanaka2020polarized}, but the most promising for mobile deployment is based on time-of-flight (ToF) imaging \cite{kirmani2009looking, velten2012recovering, o2018confocal, lindell2019wave, liu2019non, liu2020phasor, heide2014diffuse}, which extracts faint signals from hidden objects by analyzing the travel time of light. While early ToF systems were bulky and expensive \cite{velten2012recovering, velten2013femto, buttafava2015non}, recent advances in single-photon LiDAR \cite{morimoto2019megapixel, morimoto20213, liu20241, bruschini2019single, zhang2021240, kumagai20217} have made such sensors ubiquitous. Now, LiDARs are integrated into consumer mobile devices including smartphones \cite{sonyphone2021, rangwala2020}, AR headsets \cite{visionpro2024}, and autonomous vehicles \cite{sonyAV2021, ouster2020}. Although most LiDARs process raw ToF measurements into 3D point clouds of visible surfaces, the raw data also contains indirect reflections that, in principle, encode information about hidden objects.


But can consumer LiDARs be used to image hidden objects? Extensive work has been done for LiDAR-based NLOS imaging \cite{o2018confocal, ahn2019convolutional, lindell2019wave, liu2019non, liu2020phasor, liao2021fpga, xin2019theory, tsai2019beyond, young2020non, shen2021non, fujimura2023nlos, huang2023omni, chen2020learned, li2023nlost, chen2019steady}. However, these methods are designed based on idealized conditions of lab-grade LiDAR setups that do not reflect several constraints of consumer mobile sensing (\cref{tab:tech_challenges}). First, low laser power (due to eye safety constraints) and short exposure times (due to dynamic scenes) result in low signal-to-noise ratio (SNR) measurements. Second, constraints on bandwidth, power, and chip area limit spatial resolution, forcing a trade-off between aperture sampling density and total aperture size. Finally, the joint motion of the object and camera results in motion blur if improperly handled. Intuitively, these constraints result in lower-quality measurements than those typically assumed in prior work, motivating the need for algorithms specifically for the consumer NLOS regime.




In this work, we address these three challenges and demonstrate NLOS imaging on smartphone-grade LiDAR. While prior works have primarily focused on \emph{single-shot} reconstruction \cite{lindell2019wave, nam2021low}---where the signal quality within each frame is sufficient for inference---we tackle the more practical case where individual frames are insufficient in isolation, but collectively can inform a useful estimate of the hidden object. Our approach is inspired by burst photography \cite{hasinoff2016burst, ma2020quanta}, which exploits \emph{viewpoint redundancy} to improve SNR, and synthetic aperture radar \cite{sherwin1962some}, which exploits \emph{viewpoint diversity} to improve the reconstructed resolution. 

Our key technical contribution is the \textit{motion-induced aperture sampling (MAS) model}, which enables multi-frame fusion by unifying the effects of object shape, object motion, and camera motion. This model decouples the effects of motion and object shape on a frame measurement, which is conducive to efficient processing algorithms. Our work also uses the observation that many 3D vision tasks are constrained to solve for a single unknown variable: 3D scanning assumes a known camera pose and static scene geometry \cite{izadi2011kinectfusion, curless1996volumetric}; object tracking assumes a known object shape \cite{wang2019densefusion, xiang2017posecnn}; and camera localization assumes a known, static environment \cite{dellaert1999monte}. This problem formulation of solving for one of three unknowns---object shape, object position, or camera position---enables efficient and tractable inference using the MAS model.

Using these insights, we experimentally demonstrate three consumer NLOS imaging capabilities (\cref{fig:teaser}c). First, we show 3D reconstruction of static hidden objects by exploiting the natural handheld camera motion (with known pose) in mobile systems. Second, we demonstrate 3D tracking of single and multiple hidden objects of known shape. Third, we show camera localization using known static hidden objects as visual cues, which is especially useful when standard visual odometry techniques struggle (e.g. when imaging textureless or planar surfaces). We achieve each of these capabilities with real-time on-line processing, making it feasible for many practical applications.

To our knowledge, our results are the first such demonstrations of NLOS capabilities on widely-available consumer mobile LiDARs. Previously, NLOS imaging setups were restricted to expensive and bulky hardware with complicated setup and calibration procedures. Our work represents a step towards \textit{plug-and-play NLOS imaging}---we show for the first time that anyone can image hidden objects with off-the-shelf hardware ($\sim\$100$) and no challenging calibration. With the advent of consumer LiDARs, we envision our work will democratize access to NLOS imaging capabilities for practitioners across many disciplines---including robotics, computer vision, medicine, and UI/UX---and spur new applications of NLOS imaging.

\newcolumntype{C}[1]{>{\centering\arraybackslash}m{#1}}

\begin{table}[h]
\centering
\caption{Challenges with consumer non-line-of-sight imaging.}

\begin{tabular}{|C{5.5cm}|C{4.5cm}|C{4.5cm}|}
\hline
\textbf{Practical Constraints} & \textbf{Technical Challenge} & \textbf{Solution} \\
\hline
\vfill

\begin{itemize}[noitemsep, topsep=3pt, left=0em]
  \item eye-safe laser power
  \item short-exposure capture (30 Hz)
\end{itemize} 
\vfill

& low signal-to-noise ratio 
& 
\begin{itemize}[noitemsep, topsep=3pt, left=0em]
  \item burst imaging \& fusion
  \item motion \& shape priors
\end{itemize} \\
\hline
\begin{itemize}[noitemsep, topsep=3pt, parsep=0pt, left=0em]
  \item low power \& bandwidth capture
  \item chip fabrication constraints
\end{itemize} 
& aperture size vs. resolution tradeoff 
& 
\begin{itemize}[noitemsep, topsep=3pt, left=0em]
  \item synthetic aperture imaging
  \item motion \& shape priors
\end{itemize} \\
\hline
dynamic scenes & motion blur & aperture sampling model \\
\hline
\end{tabular}

\label{tab:tech_challenges}
\end{table}

\section{Results}

\subsection*{Capturing NLOS Signals with Consumer LiDAR}

In our experiments, we use a portable smartphone LiDAR system with $\sim100$ pixels, each consisting of a co-located laser emitter and single-photon avalanche diode (SPAD) sensor. Each SPAD pixel shares an optical axis with exactly one laser spot. While the configuration is similar to the confocal capture setup used by O'Toole et al. \cite{o2018confocal}, our setup is integrated onto an existing smartphone-grade LiDAR. Because all pixels and laser spots are active simultaneously, we assume that the hidden object of interest has strong retroreflective properties in order to retain the image formation model of confocal scanning setups \cite{o2018confocal}. However, we empirically show that our model works on diffuse objects as well. We assume object motion is a rigid-body translation (no rotations). The camera is handheld and free to move in an unstructured manner, but must be aimed towards the relay surface (i.e. the virtual mirror) at $z=0$, as shown in \cref{fig:teaser}a. We refer to the imaged area on the wall as the virtual aperture. The camera captures signals from the hidden object by measuring light that travels from the laser $\rightarrow$ virtual aperture $\rightarrow$ hidden object $\rightarrow$ virtual aperture $\rightarrow$ SPAD pixel. Each pixel measures picosecond-scale light transport by recording the distribution of light intensity as a function of time. From these space-time measurements, we infer properties of the hidden object. 


\subsection*{Space-Time Impulse Response}

The space-time impulse response (STIR) is a continuous-valued function $i(x, y, \tau)$ that fully describes the hidden scene. The STIR contains the temporal response $i(\tau)$ at every sampled $(x,y)$ location on the relay wall, assumed to be at $z=0$. Prior works have shown that the hidden scene can be estimated if the STIR is captured with sufficient SNR, spatio-temporal resolution, and virtual aperture size \cite{kirmani2009looking, gupta2012reconstruction}. 

Consumer LiDARs, however, capture a noisy, low-spatial-resolution, and cropped estimate of the STIR $\hat{i}(x, y, \tau)$. As a result, a single frame is not informative enough to reconstruct the hidden scene. However, if the hidden scene is static, we can leverage the portability of the LiDAR system by moving the camera and scanning the relay surface (as shown in \cref{fig:teaser}a). This camera motion results in increased synthetic aperture size and resolution due to additional spatial sampling of the STIR. Using multiple frames also improves SNR by exploiting redundancy in measurements. Once sufficiently sampled, the STIR can be used to reconstruct the hidden scene using techniques that don't rely on uniform spatial sampling \cite{velten2012recovering, liu2019non, liu2020phasor}. We show such a 3D reconstruction using filtered backprojection in \cref{fig:teaser}b.

\begin{figure}
    \centering
    \includegraphics[width=1\linewidth]{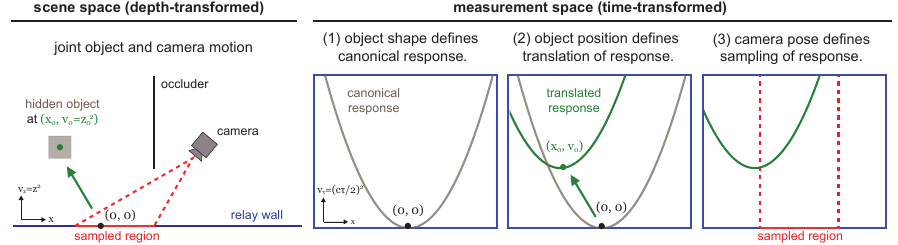}
    \caption{\textbf{Motion-induced aperture sampling model.} The space-time measurements can be expressed as the combined effect of object shape, object motion, and camera pose using the convolutional property of the light-cone transform (LCT). LCT aligns scene space and measurement space by transforming depth in scene space as $v_z=z^2$ and time in measurement space as $v_\tau=(c\tau/2)^2$.} 
    \label{fig:methods}
\end{figure}

\subsection*{Motion-Induced Aperture Sampling Model}

Under the presence of object motion, the STIR will change over time. As a result, techniques such as filtered backprojection will fail because of motion blur. For mobile applications, we must model the joint effects of object shape, object motion, and camera motion to accurately recover NLOS scenes. 

At a given object position, the light-cone transform (LCT) states that the relationship between the object's volumetric albedo $\rho(x, y, z)$ and its continuous-valued STIR $i(x, y, \tau)$ can be expressed as a 3D convolution 

\begin{equation}
    \mathcal{R}_\tau\{i\}(x, y, v) = \mathcal{R}_z \{\rho\}(x, y, v) \circledast h(x, y, v),
    \label{eq:lct}
\end{equation}

\noindent where $\mathcal{R}_\tau$ and $\mathcal{R}_z$ are non-linear resampling operators in time and space respectively, $\circledast$ is the convolution operator, and $h(x, y, v)$ is the 3D point spread function. Due to the convolutional image formation model in LCT-transformed space, we will treat all measurements in the transformed space $\mathcal{R}_\tau\{i\}(x, y, v)=\mathcal{I}(x, y, v)$. Without loss of information, we can treat $\mathcal{I}(x, y, v)$ as the STIR of the hidden scene. 

Under a rigid-body object translation of $\mathbf{\Delta}=(\Delta x, \Delta y, \Delta v)$, the STIR also translates by $\mathbf{\Delta}$ due to the shift-invariance of the convolution operation. As a result, the STIR at time $t$ can be expressed as

\begin{equation}
    \mathcal{I}_t(x, y, v)=\mathcal{I}(x-\Delta x_t, y - \Delta y_t, v - \Delta v_t).
\end{equation}

\noindent This property allows us to decompose the time-varying STIR into a canonical STIR $\mathcal{I}(x, y, v)$ that is time-independent, and an object shift $\mathbf{\Delta}_t=(\Delta x_t, \Delta y_t, \Delta v_t)$ that is time-dependent. A proof of this decomposition is provided in the \textcolor{blue}{Supplementary Material}. 

The camera acts as a spatial sampling function, sampling the STIR at spatial locations determined by the camera pose and with fixed timing resolution. The measured spatio-temporal profile at time $t$, 2D pixel location $\mathbf{u}$, and temporal bin $v$ can be expressed as 

\begin{equation}
    i_t(\mathbf{u}, v) = \iint_{A} \underbrace{\delta(\mathbf{u} - \mathbf{KP}_t\mathbf{x}')}_{\text{camera sampling}} \cdot \underbrace{\mathcal{I}(\mathbf{x} - \mathbf{\Delta}_t, v - \Delta v_t)}_{\text{object shape \& position}} d\mathbf{x},
    \label{eq:mas_model} 
\end{equation}

\noindent where $\mathbf{P}_t$ is the camera extrinsic matrix at time $t$, $\mathbf{K}$ is the camera intrinsic matrix, $\mathbf{x} = [x, y, 0]^\top$ are spatial locations on the relay wall, $\mathbf{x}'=[\mathbf{x}, 1]^\top$ is the homogeneous coordinates of $\mathbf{x}$, and $i(\mathbf{x}, v)$ is the LCT-transformed transient measurement at pixel location $\mathbf{u}$. 

\cref{eq:mas_model} defines the motion-induced aperture (MAS) sampling model. The second term encodes the time-independent object shape through the canonical STIR $\mathcal{I}(\mathbf{x}, v)$ and the time-dependent object position through $\mathbf{\Delta}_t$. The first term models the effect of the camera pose $\mathbf{P}_t$ as a spatial sampling of the continuous STIR function. In practice, the signal is also temporally quantized due to the sensor timing resolution, but we ignore it for conceptual simplicity. The combined effects of object shape, object motion, and camera motion are summarized in \cref{fig:methods}: object shape defines a canonical STIR $\mathcal{I}(x, y, v)$, object motion defines the translation $\mathbf{\Delta}$ of the canonical STIR, and camera pose $\mathbf{P}_t$ determines the sampling of the translated STIR.

\subsection*{Particle Filtering for Single and Multi-Object 3D Tracking}

\begin{figure}
    \centering
    \includegraphics[width=0.8\linewidth]{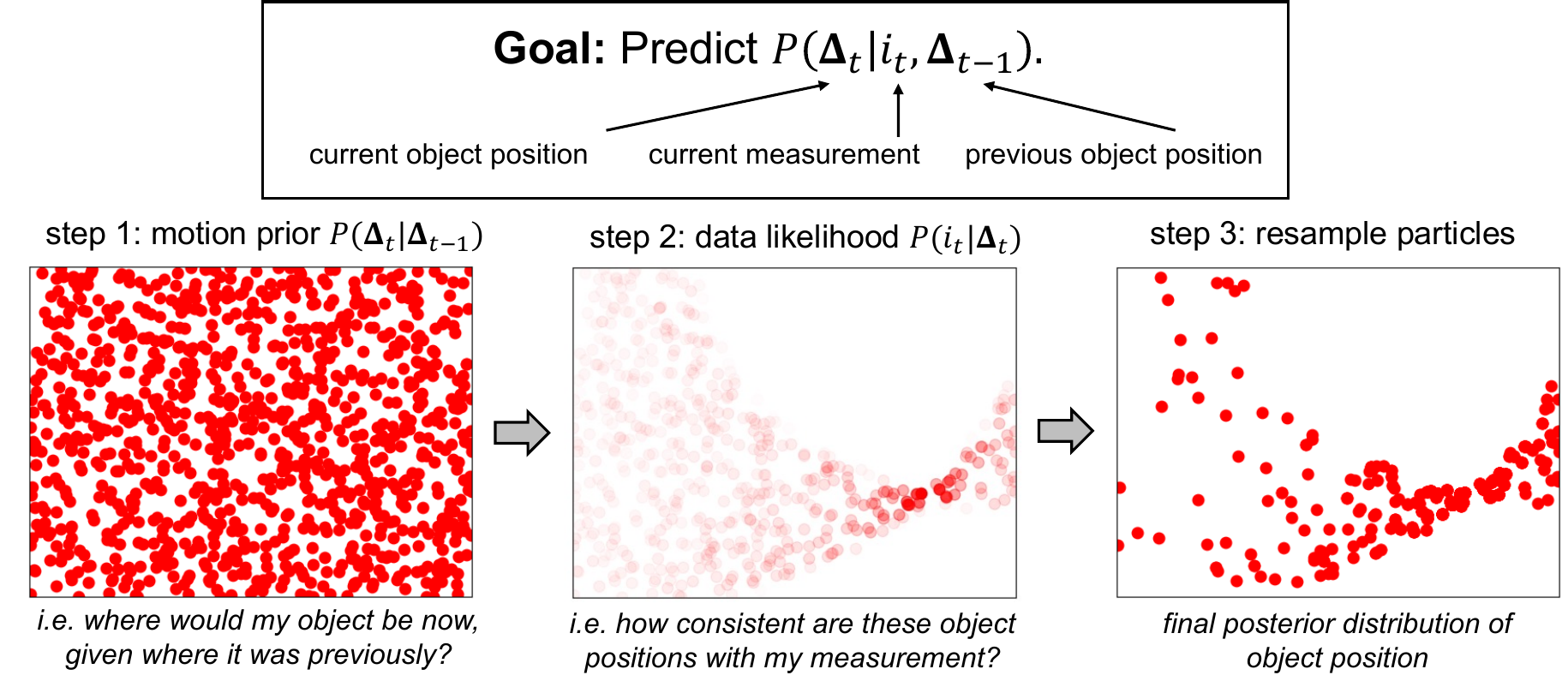}
    \caption{\textbf{Particle filtering.} Particle filtering consists of three steps: particle propagation, evaluation, and resampling. Propagation uses a motion prior and evaluation updates the prior with the data likelihood.}
    \label{fig:particle}
\end{figure}

 Jointly recovering object shape, object position, and camera pose under the MAS model is a highly non-convex inverse problem that would require iterative solvers. Such iterative solvers, however, are computationally expensive and unsuitable for real-time execution on consumer platforms. We avoid the need for such solvers by using the observation that many applications require solving for one unknown variable at a time. In tracking, the object of interest often has known (or approximately known) shape, e.g. when a LOS object moves out of view \cite{wang2019densefusion, xiang2017posecnn}. In robotics, camera localization assumes that the environment is known when recovering the camera position \cite{dellaert1999monte}. Motivated by such applications, we outline a technique for real-time object tracking and camera localization based on the MAS model. We first focus on object tracking and later extend it to camera localization.


NLOS tracking is challenging because of the low spatio-temporal resolution and low SNR of the measurements. As a result, we want to be able to (a) robustify our estimate of the object position by incorporating information from past measurements, and (b) model the uncertainty of our estimates based on the quality of the measurements. We employ a particle filter to achieve both of these goals.

Our goal is to model the posterior distribution $P(\mathbf{\Delta}_t \mid i_t, \mathbf{\Delta}_{t-1})$, i.e., the probability of the current object position $\mathbf{\Delta}_t$ given the current measurement $i_t$ and previous object position $\mathbf{\Delta}_{t-1}$. A particle filter allows us to represent arbitrarily complex distributions by exploiting the duality between a distribution and samples generated under it \cite{smith1992bayesian, dellaert1999monte}. As a result, we can model the posterior by storing $N$ random samples (i.e. particles) from the distribution. The $k$th particle $\mathbf{\Delta}_{t,k} \in \mathbb{R}^d$ represents a guess of the object position at time $t$, where $d=3$ because the state is the 3D position of a single object. The final posterior distribution can be estimated using kernel density estimation \cite{scott2015multivariate}.

The particle filtering algorithm enables sequential processing through a decomposition of the posterior distribution into two terms: the motion model and the data likelihood

\begin{equation}
    \underbrace{P(\mathbf{\Delta}_t \mid i_t, \mathbf{\Delta}_{t-1})}_{\text{posterior}} \propto \underbrace{P(\mathbf{\Delta}_t \mid \mathbf{\Delta}_{t-1})}_{\text{motion model}} \cdot \underbrace{P(i_t \mid \mathbf{\Delta}_t)}_{\text{data likelihood}}.
\end{equation}

\noindent This decomposition enables straightforward estimation of the posterior through three steps: particle propagation, evaluation, and resampling. First, each particle is propagated based on the probabilistic motion model $P(\mathbf{\Delta}_t \mid \mathbf{\Delta}_{t-1})$. Choice of the motion model builds in a prior as to how the object moves from one time step to the next (e.g. constant velocity). Second, each particle is evaluated using a score function that defines the data likelihood term $P(i_t \mid \mathbf{\Delta}_t)$. This step assigns a weight to each particle by (a) rendering the measurement via the MAS model and (b) assigning a score based on how similar the rendering is to the measurement. Finally, the particles are resampled using importance sampling. Intuitively, the resampling step probabilistically eliminates particles that are highly inconsistent with the measurement and introduces particles that are more consistent with the measurement. After this step, all particles have equal weight and collectively parametrize the posterior. The particle filter algorithm is summarized in \cref{fig:particle} and \cref{alg:tracking}.

We demonstrate single-object tracking in \cref{fig:teaser}b. For these results, we omit plotting the first few frames, which correspond to the object localization stage. We use the aforementioned particle filtering technique to estimate the object position. In order to output a single position estimate for each frame (rather than distribution), we compute the mean of the distribution by computing the mean of the particles. The trajectory is then plotted as the mean estimate over all frames. The predicted trajectory is shown in blue and the ground truth trajectory is in black. NLOS imaging is inherently ill-posed, and many object positions will be consistent with the measurements. A useful feature of particle filtering is that it can implicitly capture these regions of ambiguity through the posterior, and even break the ill-posedness to an extent by leveraging the motion prior. We also compare our proposed approach for tracking to approaches based on conventional backprojection techniques \cite{chan2017non, gariepy2015tracking} in the \textcolor{blue}{Supplementary Material}.

\begin{figure}
    \centering
    \includegraphics[width=1\linewidth]{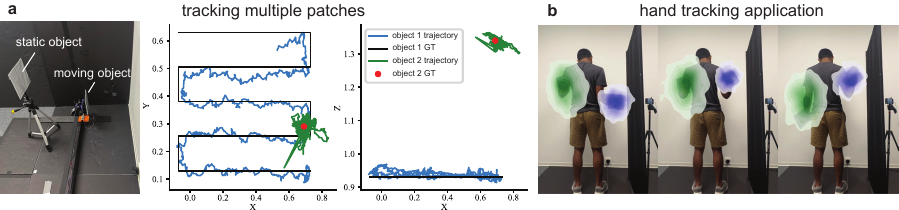}
    \caption{\textbf{Tracking multiple hidden objects.} (a) We can detect and track multiple objects. In this result, we show tracking of one moving object on a translation stage and one static object. The first few frames are localizing the object, and are therefore omitted from the visualization. (b) We demonstrate NLOS hand tracking as an application of multi-object tracking. The predicted right and left hand distributions are shown in blue and green respectively.}
    \label{fig:multi-object}
\end{figure}

We can also perform multi-object tracking in a similar manner with a small modification to the particle filtering formulation. If we want to track $M$ objects, then the size of our state space would be $d=3M$ and the particle filter would represent the joint probability distribution of all object positions 

\begin{equation}
    P(\mathbf{\Delta}^{(1)}_t, ..., \mathbf{\Delta}^{(M)}_t \mid i_t, \mathbf{\Delta}^{(1)}_{t-1}, ..., \mathbf{\Delta}^{(M)}_{t-1}),
\end{equation}

\noindent where $\mathbf{\Delta}_t^{(m)}$ represents the position of the $m$th object at time $t$. The rendered measurement of the $k$th particle at time $t$ would then be the linear super-position of all object renderings

\begin{equation}
    \hat{i}_{t,k}(\mathbf{u}, v) = \sum_{m=1}^M w_{t, k} \cdot\hat{i}_{t,k}^{(m)}(\mathbf{u}, v), 
\end{equation}

\noindent where $w_{t,k}$ is the dot product between $\hat{i}^{(m)}_{t,k}/\vert \hat{i}^{(m)}_{t,k} \vert_2$ and the measurement $i_{t,k}$, and $\hat{i}^{(m)}_{t,k}$ is the rendered contribution of the $m$th object. We found that applying this weight normalization and ensuring that $\mathcal{I}_0$ accounts for laser pulse width were crucial for performance. In addition, simply computing the mean of the posterior distribution does not work for multi-object tracking because the distribution is multi-modal (due to multiple objects). Instead, we cluster particles using $K$-means clustering, choose the cluster with the most amount of particles, then estimate the positions of the objects by computing the mean within that cluster. While we do not account for object occlusions here, the model could easily account for them by modifying \cref{eq:mas_model} to include a pixel-wise visibility term that models the visibility between pixel $\mathbf{u}$ and object $m$ based on other objects. We demonstrate multi-object tracking in \cref{fig:multi-object}. We also demonstrate NLOS hand tracking as an application enabled by our method. The probability distribution of the left and right hand position are shown in green and blue contour maps respectively. Because the left hand is further from the virtual aperture than the right hand, it's position is more uncertain due to a larger region of ambiguity. 


\subsection*{Camera Localization using NLOS Objects}

We can use signals from hidden objects to localize the camera, especially when line-of-sight surfaces do not contain useful visual features (e.g. planar white surfaces). Conventional visual odometry techniques (e.g. SfM, ICP) would fail in such scenarios because of limited geometric and texture variation. However, if we can image outside the camera's line-of-sight, we can use nearby objects as landmarks.

The goal of camera localization is to recover the position of the camera in a defined world coordinates. Because the hidden object serves as the landmark, we define the origin of this coordinate system to be at the same $(x, y)$ location as the object, with $z=0$ being the wall plane. In the case where we are imaging a planar surface, point cloud measurements from a LiDAR would constrain the 6D camera pose up to three degrees of ambiguity: 2D translation parallel to the wall and roll rotation parallel to the wall. Assuming no roll rotation parallel to the wall, only the $(x, y)$ translation component of the pose $\mathbf{P}_t$ is unknown. Mathematically, we could express the world coordinates of a point $\mathbf{x}$ on the wall imaged by pixel $\mathbf{u}$ as

\begin{align}
    \mathbf{x} &= \mathbf{P}^{-1}\mathbf{K}^{-1}\mathbf{u} \\
    &= [\mathbf{R} \mid \mathbf{t}] \cdot \mathbf{K}^{-1}\mathbf{u} \\
    &= \underbrace{\mathbf{R}}_{\text{estimated}} \cdot \underbrace{\mathbf{K}^{-1}\mathbf{u}}_{\text{measured}} + \underbrace{\mathbf{t}}_{\text{unknown}}.
\end{align}

\noindent Here, $\mathbf{P}^{-1}$ is the unknown cam2world matrix that we can decompose into a rotation $\mathbf{R}$ and translation $\mathbf{t}$ component. The camera coordinates of the imaged point $\mathbf{x}_c=\mathbf{K}^{-1}\mathbf{u}$ is known from the LiDAR depth and the camera intrinsics $\mathbf{K}$. If there is no roll rotation with respect to the wall, then $\mathbf{R}$ can also be estimated via plane fitting. The $z$ location of the camera can then deterministically be computed from the fitted point cloud $\hat{\mathbf{x}}_c$. Only the $(x, y)$ component of the cam2world translation vector $\mathbf{t} = [x_{c,t}, y_{c,t}, z_{c, t}, 1]$ is unknown. 

We can estimate this translation component by using the NLOS signal. The true point cloud is $\mathbf{x}_w=\hat{\mathbf{x}}_c + [t_x, t_y, 0]^\top$. Using \cref{eq:mas_model}, we can render what the measurement looks like under different translations $[t_x, t_y]$, assuming the object shape (i.e. the canonical STIR) is known. This problem formulation naturally allow us to integrate the camera localization problem into a particle filtering framework similar to the object tracking case. The camera localization algorithm is summarized in \cref{alg:localization}. We show quantitative results in \cref{fig:teaser}b and qualitative results with unstructured handheld motion in the \textcolor{blue}{Supplementary Video}.


\begin{figure}
    \centering
    \includegraphics[width=1\linewidth]{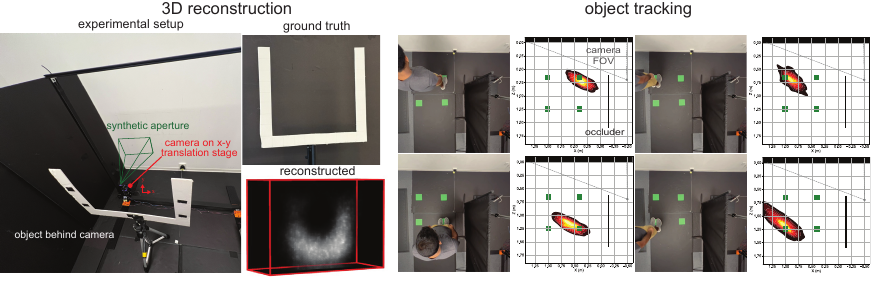}
    \caption{\textbf{Imaging hidden diffuse objects.} Although the MAS model we propose was derived assuming retroreflective object reflectance, we find empirically that the model can handle diffuse objects. We show that we are still able to reconstruct 3D shape and track objects in real-time (available as supplementary video). The results, however, are inherently worse than those obtained with retroreflective objects because of the weaker signal (due to $r^4$ falloff) and light contributions from non-confocal paths.}
    \label{fig:diffuse}
\end{figure}

\section{Discussion}

The primary goal of this paper is to demonstrate the feasibility of many non-line-of-sight imaging capabilities using already-widely-available consumer mobile LiDARs. Specifically, we show that these sensors can be used to image hidden objects for three applications: (1) 3D object reconstruction, (2) single- and multi-object 3D tracking, and (3) camera localization using NLOS objects as visual cues. While most prior works have shown proof-of-concept results using lab-based hardware setups, our work directly addresses the challenges associated with consumer sensors: resolution, SNR, aperture size, and motion.

We show that our proposed motion-induced aperture sampling can handle 6D camera motion when pose is known and 5D camera motion when pose is unknown. While we don't show results for 6D motion in camera localization, our model can handle the extra degree of motion as long as the object shape is not rotationally symmetric. The MAS model can also handle 4D object motion (3D translation and 1D roll rotation parallel to wall), but we only show the translation case. Future work could handle non-rigid-body motion by modeling a point-wise deformation field \cite{park2021nerfies, pumarola2021d} of the canonical STIR. Handling object pitch and yaw rotation would require modeling visibility of points that come in and out of view of the relay surface.

We are able to handle challenges related to SNR and reconstruction resolution in tracking and camera localization by leveraging multi-frame fusion. Specifically, we leverage particle filtering to improve SNR by exploiting motion priors and handle resolution limits by explicitly modeling regions of ambiguity using probability distributions. One limitation of our work is that we do not model much of the complexity of the real world, such as complex reflectance properties of natural objects (e.g. inhomogeneous BRDFs), the viewing angle of the camera with respect to the relay surface, and sensor non-idealities (e.g. pulse width). As a result, it is challenging to derive a score function that can robustly compare the rendered and actual measurement. In particular, we notice larger regions of ambiguity when dealing with larger objects. Future work could overcome this limitation by leveraging machine learning to learn a score function, rather than relying on handcrafted score functions. 

Our model assumed that most measured light resulted from retroreflective reflections. Experimentally, these conditions held because we used objects covered in retroreflective materials. We also demonstrate that our model handles diffuse objects well in practice, as shown in \cref{fig:diffuse}. However, the signal quality is worse due to poorer SNR and stronger reflections from non-confocal paths. 


In this work, we assumed that two out of object shape, object motion (static or dynamic), and camera pose are known. Future work could extend this assumption to directly solving for all three unknowns, similar to the SLAM formulation \cite{montemerlo2002fastslam}. However, solving for object motion and camera motion jointly can be challenging due to degenerate cases, e.g., if the object and camera translate the same amount parallel to the $x-y$ plane. In such cases, only the relative motion of the object and camera can be recovered.



While our work is focused on non-line-of-sight imaging, we believe that there are other practical uses of multi-bounce signals in consumer LiDARs. One such example is 3D mapping in constrained setting where the number of pixels is limited (e.g. minimalist sensing \cite{klotz2024minimalist}). In such cases, 3D mapping is challenging because conventional LiDAR maps a single scene point to a single point in the measurement. When fewer pixels are available, the measurement of the 3D shape will be aliased due to low spatial sampling rate. On the other hand, the third-bounce measurements of an object results in a blurring of the object shape with a PSF kernel $h(x, y, v)$. Sampling this blurred signal is a form of anti-aliasing that can be useful for recovering the rough shape of an object. This idea is further discussed in the \textcolor{blue}{Supplementary Materials}. 

Historically, NLOS imaging has required expensive equipment with challenging calibration procedures. However, with the low-cost commercial availability of single-photon LiDAR sensors, we envision a new era of \emph{plug-and-play NLOS imaging}. Instead of requiring tens of thousands of dollars and hours of calibration, plug-and-play NLOS imaging enables NLOS capabilities with $<\$100$ and seconds of calibration. In \textcolor{blue}{Supplementary Materials}, we show that a commercially-available ST VL53L8CX \cite{stspad} can be used for NLOS tracking with no physical calibration or additional hardware needed. The code required to run such demonstrations will be publicly released. We hope that such plug-and-play capabilities will democratize access to NLOS imaging and spur further research along this direction.

\section{Methods}

\subsection*{Pre-Computing Canonical STIR and Rendering as an Indexing Operation}

In \cref{eq:mas_model}, we showed that the object shape can be fully described by the continuous space-time impulse response (STIR) of the object. The relationship between the object shape and this canonical STIR is determined by the PSF kernel

\begin{equation}
    h(x, y, v) = \delta(x^2 + y^2 - v),
\end{equation}

\noindent which parametrizes a parabola in LCT-transformed space-time. As a result, the canonical STIR of the object can be approximate as a sum of parabolas

\begin{equation}
    \mathcal{I}(x, y, v) \approx \sum_{p \in \mathcal{P}} \rho_p \cdot h(x-p_x, y-p_y,  v-p_v),
\end{equation}

\noindent where $\mathcal{P}$ is the point cloud defining the object shape, $p=(p_x, p_y, p_z)$, $p_v=p_z^2$, and $\rho_p$ is the albedo at the $p$. We normalize the point cloud such that the centroid of the points is defined to be at the origin. We assume that the object has uniform albedo and set $\rho_p=1$. 

Under object translation, the object shape as viewed by the relay surface is time-invariant. As a result, we can pre-compute the canonical measurement $\mathcal{I}$ as a voxel cube $\hat{\mathcal{I}} \in \mathbb{R}^{N_x \times N_y \times N_v}$, where $N_x$, $N_y$, and $N_v$ are the number of points sampled along the $x$, $y$, and $v$ direction of the canonical STIR. Once this STIR is pre-computed at pre-determined spatial-temporal points, any measurement can be efficiently rendered by simply indexing into the correct space-time points using \cref{eq:mas_model}.

\subsection*{Details on Particle Filtering}

We use $1000$ particles to parametrize the posterior distribution. The three main steps of particle filtering are particle propagation, evaluation, and resampling. Particle propagation requires specification of a motion model or prior, which we define to be 

\begin{equation}
    \mathbf{\Delta}_{t} \sim \mathcal{N}(\mathbf{\Delta}_{t-1}, r\mathbf{I}), 
\end{equation}

\noindent where $\mathbf{I}$ is the identity matrix, $\mathcal{N}(\mu, \sigma)$ is a normal distribution with mean $\mathbf{\mu}$ and covariance $\mathbf{\Sigma}$, and $r$ is a measure of the maximum expected radius that the object is expected to travel between consecutive frames. The motion prior can also be based on second-order information (e.g. velocity, acceleration) but we find that a proximity prior with $r=5$ cm already works well empirically at $30$ Hz. The particle evaluation step requires comparison of the rendered measurement $\hat{i}_{t, k} \in \mathbb{R}^{n_x \times n_y \times n_v}$ of the $k$th particle with the actual measurement $i_t \in \mathbb{R}^{n_x \times n_y \times n_t}$. We use a normalized dot product score function

\begin{equation}
    \text{score}_k =  \left[ \frac{i_t \odot \hat{i}_{t, k}}{\vert i_t \vert \cdot\vert \hat{i}_{t, k}\vert} \right] ^ \eta,
    \label{eq:score-function}
\end{equation}

\noindent where $\text{score}_k$ is the unnormalized score of the $k$th particle, and $\eta$ is a hyperparameter governing the relative strength of the motion prior and the data likelihood. Intuitively, a higher value of $\eta$ increases the separation between lower-valued and higher-valued particles, indicating that the algorithm should rely on the data likelihood more than the motion prior. When $\eta$ is set to be lower, the algorithm will not favor one particle significantly more than the other, reducing the impact of the data likelihood. This case is useful when individual measurements are less reliable or more noisy. While $\eta$ could be defined adaptively (e.g. based on per-frame SNR), we set $\eta$ to be fixed within an experiment. Once the score is computed for all particles, the score is normalized to produce a valid probability distribution. Finally, we use residual resampling \cite{hol2006resampling} to maintain a balance between low-variance sampling and particle diversity. The pseudocode algorithm for object tracking and camera localization are provided in \cref{alg:tracking} and \cref{alg:localization} respectively.

\begin{algorithm}
\caption{Single and Multi-Object Tracking with Known Pose and Object Shape(s)}
\begin{algorithmic}[1]
\State \textbf{Input:} number of particles $K$, space-time measurements $i_{1:T} \in \mathbb{R}^{n_xn_y \times n_v}$, point clouds in world coordinates $p_{1:T} \in \mathbb{R}^{n_xn_y \times 3}$, canonical STIR of $M$ hidden objects $\{\hat{\mathcal{I}}_1(x, y, v),...,\hat{\mathcal{I}}_M(x, y, v)\} \in \mathbb{R}^{N_x \times N_y \times N_v}$
\State \textbf{Output:} posterior distribution over $(x, y, z)$ object positions $\mathbf{\Delta}_{1:T} \in \mathbb{R}^{K \times 3M}$

\State

\State \textbf{\# === Initialize particles at $t=0$ from $3M$-D uniform distribution === \#}


\State $\{\{\mathbf{\Delta}_{t=1,k,m}\}_{k=1}^K\}_{m=1}^M \gets \mathcal{U}\left([x_{\min}, x_{\max}] \times [y_{\min}, y_{\max}] \times [z_{\min}, z_{\max}] \right) $ \Comment{$K \times 3M$ uniform matrix}

\State

\State \textbf{\# === Loop through frames === \#}
\For{$t = 1$ to $T$}    
    \State \textbf{\# Transform measurements using LCT}
    \State $i_t \gets \text{transform\_measurements}(i_t)$

    \State
    \State \textbf{\# Evaluate particles (data likelihood)}
    \For{$k = 1$ to $K$}
        \State initialize $\hat{i}_{t,k} \gets 0$
        \For{$m = 1$ to $M$}
            \State $\hat{i}_{t,k} \gets \hat{i}_{t,k} + \text{render}(\hat{\mathcal{I}}_m, p_t, \mathbf{\Delta}_{m,k})$ \Comment{Eq. 3}
        \EndFor
        \State $w_{t,k} \gets \text{dot\_product\_score}(i_t, \hat{i}_{t,k})$  \Comment{Eq. 10}
    \EndFor

    \State  $w_{t,k} \gets \frac{w_{t,k}}{\sum_{j=1}^N w_{t,j}}$ \Comment{normalize particle scores}

    \State

    \State \textbf{\# Resample particles based on particle weights}
    \State 
    $\{\mathbf{\Delta}_{t,k}\}_{k=1}^K \gets \text{residual\_resampling}(\{\mathbf{\Delta}_{t,k}, w_{t,k}\}_{k=1}^K)$

    \State
    \State \textbf{\# Propagate particles for next frame (motion model)}
    \State $\mathbf{\Delta}_{t+1,k} \gets \mathcal{N}(\mathbf{\Delta}_{t,k}, r\mathbf{I})$
    
\EndFor
\end{algorithmic}
\label{alg:tracking}

\end{algorithm}
\begin{algorithm}
\caption{Camera Localization using Static Hidden Object}
\begin{algorithmic}[1]

\State \textbf{Input:} number of particles $K$, space-time measurements $i_{1:T} \in \mathbb{R}^{n_xn_y \times n_v}$, point clouds in camera coordinates $p_{1:T} \in \mathbb{R}^{n_xn_y \times 3}$, canonical STIR of hidden object $\hat{\mathcal{I}}(x, y, v) \in \mathbb{R}^{N_x \times N_y \times N_v}$
\State \textbf{Output:} posterior distribution over $(x, y)$ camera locations $\mathbf{c}_{1:T} \in \mathbb{R}^{K \times 2}$, camera $z$ positions $z_{1:T}$

\State

\State \textbf{\# === Initialize particles at $t=1$ from 2D uniform distribution === \#}

\State $\{\mathbf{c}_{t=1,k}\}_{k=1}^K \gets \mathcal{U}\left([x_{\min}, x_{\max}] \times [y_{\min}, y_{\max}]\right) $ \Comment{$K \times 2$ uniform matrix}

\State

\State \textbf{\# === Loop through frames === \#}
\For{$t = 1$ to $T$}
    \State \textbf{\# Fit point cloud to $z=0$ to compute point cloud and camera $z$ position}
    \State $z_t, p_t \gets \text{fit\_plane}(p_t)$ 

    \State
    
    \State \textbf{\# Evaluate particles (data likelihood)}
    \For{$k = 1$ to $K$}
        \State $p_{t, k}[:, 0] \gets p_t[:, 0] + \mathbf{c}_k[0]$ \Comment{$x$ point cloud shift}
         \State $p_{t, k}[:, 1] \gets p_t[:, 1] + \mathbf{c}_k[1]$ \Comment{$y$ point cloud shift}
        
        \State $\hat{i}_{t,k} \gets \text{render}(\hat{\mathcal{I}}, p_{t,k}, 0)$ \Comment{render with no object shift (Eq. 3)}
        \State $w_k \gets \text{dot\_product\_score}(i_t, \hat{i}_{t,k})$  \Comment{score function (Eq. 10)}
    \EndFor

    \State  $w_{t,k} \gets \frac{w_{t,k}}{\sum_{j=1}^N w_{t,j}}$ \Comment{normalize particle scores}

    \State

    \State \textbf{\# Resample particles based on particle weights}
    \State 
    $\{\mathbf{c}_{t,k}\}_{k=1}^K \gets \text{residual\_resampling}(\{\mathbf{c}_{t,k}, w_{t,k}\}_{k=1}^K)$

    \State
    \State \textbf{\# Propagate particles for next frame (motion model)}
    \State $\{\mathbf{c}_{t+1,k}\}_{k=1}^K \gets \mathcal{N}(\mathbf{c}_{t,k}, r\mathbf{I})$
    
\EndFor
\end{algorithmic}
\label{alg:localization}

\end{algorithm}

\bibliographystyle{ieeetr}
\bibliography{bib}

\clearpage

\def\maketitle{%
  \begin{center}
    {\LARGE\bfseries Supplementary: Imaging Hidden Objects with Consumer LiDAR via Motion Induced Sampling\par}
    \vspace{1em}
    {\large Siddharth Somasundaram$^1$, Aaron Young$^1$, \\ Akshat Dave$^1$, Adithya Pediredla$^2$, Ramesh Raskar$^1$\par}
    \vspace{0.5em}
    {\footnotesize $^1$Massachusetts Institute of Technology, Cambridge, MA, United States \\ $^2$Dartmouth College, Hanover, NH, United States\par}
    \vspace{1em}
    \fbox{%
      \begin{minipage}{0.52\linewidth}
        \centering
        {\large Project Page:}
        \href{https://sidsoma.com/consumer-nlos/}{\large sidsoma.com/consumer-nlos/}
      \end{minipage}
    }
    \vspace{1em}
  \end{center}
}
\setcounter{section}{0}
\renewcommand{\thesection}{S\arabic{section}}

\date{}
\maketitle
\thispagestyle{fancy}

\section{Derivation of Motion-Induced Aperture Sampling Model}

\paragraph{Primer on Light-Cone Transform.}
Our NLOS setup consists of a planar relay wall that lies along the $z=0$ plane. Under a confocal setup with a static object, the measured intensity at a point on the wall $(x, y, z=0)$ and time $\tau$ can be expressed as 

\begin{equation}
    i(x,y,\tau) = \iiint_\Omega \frac{\rho(x', y', z')}{r(x',y',z')^4}\cdot \delta(2r(x', y', z')-c\tau)dx'dy'dz',
    \label{eq:cnlos_forward_model}
\end{equation}

\noindent where $\rho(x', y', z')$ is the volumetric albedo of the hidden scene, $r(x', y', z') =\sqrt{(x-x')^2 + (y-y')^2 + {z'}^2}$, $c$ is the speed of light, $\delta(\cdot)$ is the Dirac delta function, and the integral is over the 3D hidden volume. This model assumes diffuse reflectance, but can also be modified to account for retroreflective objects by changing the $r^4$ term to $r^2$. After applying a change of variables $z'=\sqrt{u}$ and $v=(ct/2)^2$, noting that $r(x', y', z') = (ct/2) = \sqrt{v}$, and using properties of Dirac delta functions, \cref{eq:cnlos_forward_model} becomes

\begin{equation}
    \underbrace{v^{3/2} \tau(x, y, \frac{2\sqrt{v}}{c})}_{\mathcal{R}_t\{\tau\}(x, y, v)} = \iiint_\Omega \underbrace{\frac{\rho(x', y', \sqrt{u})}{2\sqrt{u}}}_{\mathcal{R}_z\{\rho\}(x, y, u)} \cdot \underbrace{\delta((x-x')^2+(y-y')^2+u-v)}_{h(x-x', y-y', v-u)}dx'dy'du,
    \label{eq:lct_full}
\end{equation}

\noindent where $\mathcal{R}_t$ and $\mathcal{R}_z$ are non-linear resampling and attenuation operators along $t$ in the measurement space and $z$ in the albedo space. $\mathcal{R}_t$ transforms the time axis of the measurement as $v=(ct/2)^2$ and scales the measurement intensity by $v^{3/2}=(ct/2)^3$. $\mathcal{R}_z$ transforms the depth axis of the hidden volume as $u=z^2$ and scales the albedo by $1/2\sqrt{u}=1/2z$. \cref{eq:lct_full} defines the light-cone transform (LCT) \cite{o2018confocal}, and has convolutional form in a transformed measurement and object space

\begin{align}
    \mathcal{I}(x, y, v) =  \mathcal{Q}(x, y, v) \circledast h(x, y, v),
    \label{eq:lct}
\end{align}

\noindent where $\mathcal{Q} = \mathcal{R}_z\{\rho\}$ and $\mathcal{I} = \mathcal{R}_t\{i\}$. The LCT performs a 3D convolution in $x$, $y$, and the transformed variable $v$. $v$ refers to the transformed version of the $t$ axis of the measurement $i$ and the $z$ axis of the volumetric albedo $\rho$. In transforming these two dimensions, $t$ and $z$, the LCT aligns time and space to make 3D convolution mathematically coherent. We leverage this convolutional relationship between the measurements $\tau$ and object $\rho$ in the LCT-transformed space to develop our motion model. 



\paragraph{Effect of Object Motion.}
\label{sec:object_motion}
We now consider the effect that object translation\footnote{We don't consider rotation in this work, though our model could be extended to roll rotation parallel to the wall.} has on the space-time impulse response $\mathcal{I}(x, y, v)$. An object with shape $\rho(x, y, z) \xLeftrightarrow{\text{LCT}} Q(x, y, v)$ has an impulse response $\mathcal{I}_0(x, y, v)$ as defined by \cref{eq:lct}. We refer to $\mathcal{I}_0(x, y, v)$ as the canonical response of the object.

We approximate the object's shape as a set of points $\mathcal{P}$, where each point $p$ has an albedo $\rho_p$. Therefore, the object shape can be modeled as 

\begin{equation}
    \rho(\mathbf{x}_p, z) \approx \sum_{p \in \mathcal{P}} \rho_p \cdot \delta(\mathbf{x}-\mathbf{x}_p, z-z_p),
\end{equation}

\noindent where $\mathbf{x}=(x, y)$ and $\mathbf{x}_p=(x_p, y_p)$. Using this parametrization of the object shape and neglecting self-occlusions, the impulse response in LCT space is

\begin{equation}
    \mathcal{I}_0(\mathbf{x}, v)= \left[ \sum_{p \in \mathcal{P}} \frac{\rho_p}{2\sqrt{v_p}} \cdot \delta(\mathbf{x}-\mathbf{x}_p, v-v_p)\right] \circledast h(\mathbf{x}, v),
\end{equation}

\noindent where $v_p = z_p^2$. Assuming a rigid-body object translation of $\mathbf{\Delta}=(\Delta_x, \Delta_y, \Delta_z)$, each point on the object will move by the same amount $\mathbf{\Delta}$. As a result, the impulse response after object translation can be expressed as 

\begin{align}
    \mathcal{I}(\mathbf{x}, v) &= \left[ \sum_{p \in \mathcal{P}} \frac{\rho_p}{2\sqrt{v_p + \Delta_v}} \cdot \delta(\mathbf{x}-\mathbf{x}_p-\mathbf{\Delta}_x, v-v_p-\Delta_v) \right] \circledast h(\mathbf{x}, v) \\
    &\approx \left[ \sum_{p \in \mathcal{P}} \frac{\rho_p}{2\sqrt{v_p}} \cdot \delta(\mathbf{x}-\mathbf{x}_p-\mathbf{\Delta}_x, v-v_p-\Delta_v) \right] \circledast h(\mathbf{x}, v) \\
    &= \left[ \sum_{p \in \mathcal{P}} \frac{\rho_p}{2\sqrt{v_p}} \cdot \delta(\mathbf{x}-\mathbf{x}_p, v-v_p) \circledast \delta(\mathbf{x}-\mathbf{\Delta}_x, v-\Delta_v) \right] \circledast h(\mathbf{x}, v) \\
    &= \left[ \sum_{p \in \mathcal{P}} \frac{\rho_p}{2\sqrt{v_p}} \cdot \delta(\mathbf{x}-\mathbf{x}_p, v-v_p) \right] \circledast \delta(\mathbf{x}-\mathbf{\Delta}_x, v-\Delta_v) \circledast h(\mathbf{x}, v) \\
    \label{eq:tracking_as_convolution}
    &= \mathcal{I}_0(\mathbf{x}, v) \circledast \delta(\mathbf{x}-\mathbf{\Delta}_x, v-\Delta_v)  \\
    &= \mathcal{I}_0(\mathbf{x}-\mathbf{\Delta}_x, v-\Delta_v), 
\end{align}

\noindent where the second line assumes $\Delta_v << v_p$ and fourth line uses the linearity of the convolution operation. While the assumption in the second line is often not accurate, it only results in a radiometric distortion and has no effect on the shape of the space-time measurement. Therefore, we ignore it. Based on this derivation, the premise is straightforward: a shift in the object position results in a shift of the canonical measurement $\mathcal{I}_0$. The canonical response defines the impulse response at a reference object position (ideally the world coordinates origin). Any object position can be expressed with respect to the reference position defined by the canonical response. If the reference position is at $z=0$, then $\Delta_v$ would correspond to the distance of the object from the wall, and we can directly compute the $z$ displacement as $z=\sqrt{\Delta_v}$.

\paragraph{Effect of Pulse Width.} A key difference between consumer LiDARs and research-grade LiDARs is that research-grade LiDARs have much narrower pulse widths. As a result, they can be reasonably well-approximated by delta functions, as modeled in \cref{eq:cnlos_forward_model}. However, consumer LiDARs often have much wider pulse widths than research-grade LiDARs, suggesting the importance of understanding it's role in the confocal image formation model. Suppose that our laser pulse can be modeled as an unnormalized temporal Gaussian function

\begin{equation}
    f(t; \mu,\sigma) = \exp\left(\frac{-(t-\mu)^2}{2\sigma^2}\right),
\end{equation}

\noindent where $\mu$ is the time of flight and $\sigma$ is the standard deviation governing the pulse width. Then, the confocal image formation model can be expressed as 

\begin{equation}
     i(x, y, t) = \iiint_\Omega \frac{1}{r^4} \cdot \rho(x', y', z')\cdot f(t; \mu(x, y; x', y', z'), \sigma)dx' dy' dz',
    \label{eq:image-formation}
\end{equation}

\noindent where $\mu(x, y; x', y', z') = \frac{2\sqrt{(x-x')^2+(y-y')^2+z'^2}}{c}$ and $c$ is the speed of light. We can rewrite the term $f(t; \mu(x, y; x', y', z'), \sigma)$ as 

\begin{align}
    f(t; \mu(x, y; x', y', z'), \sigma) &= \exp \left(\frac{-(t-\frac{2\sqrt{(x-x')^2+(y-y')^2+z'^2}}{c})^2}{2\sigma^2}\right) \\
    &= \exp\left(\frac{-(ct-2\sqrt{(x-x')^2+(y-y')^2+z'^2})^2}{2c^2\sigma^2}\right) \\
    &= \exp\left(\frac{-[(ct)^2 - 4((x-x')^2+(y-y')^2+z'^2)]^2}{2c^2\sigma^2 \cdot (ct+2\sqrt{(x-x')^2+(y-y')^2+z'^2})^2}\right) \\
    &\approx \exp\left(\frac{-[(ct)^2 - 4((x-x')^2+(y-y')^2+z'^2)]^2}{2c^2\sigma^2 \cdot 4\sqrt{(x-x')^2+(y-y')^2+z'^2})^2}\right) \\
    &= \exp\left(\frac{-[(\frac{ct}{2})^2 - ((x-x')^2+(y-y')^2+z'^2)]^2}{2c^2\sigma^2 \cdot [(x-x')^2+(y-y')^2+z'^2]}\right).
\end{align}

\noindent The third line multiplies the numerator and denominator by $(ct+2\sqrt{(x-x')^2+(y-y')^2+z'^2})^2$, and the fourth line approximates $ct \approx 2\sqrt{(x-x')^2 + (y-y')^2+z'^2}$. This term is usually true for delta functions, but is approximately true for pulses with finite pulse width and is bounded by $\sigma$. Plugging this expression into \cref{eq:image-formation} yields

\begin{equation}
     i(x, y, t) = \iiint_\Omega \frac{1}{r^4} \cdot \rho(x', y', z')\cdot  \exp\left(\frac{-[(\frac{ct}{2})^2 - ((x-x')^2+(y-y')^2+z'^2)]^2}{2c^2\sigma^2 \cdot [(x-x')^2+(y-y')^2+z'^2]}\right) dx' dy' dz'
\end{equation}

\noindent Substituting $v=(\frac{ct}{2})^2$, $z = \sqrt{u}$, and noting that $r=\frac{ct}{2}=\sqrt{v}$, we are left with

\begin{align}
    v^2 \cdot i(x', y', \frac{2\sqrt{v}}{c}) &= \iiint_\Omega \frac{1}{2\sqrt{u}}\rho(x', y', \sqrt{u})\cdot \underbrace{\exp \left( \frac{-[v-((x-x')^2+(y-y')^2+u)]^2}{2c^2\sigma^2 \cdot [(x-x')^2+(y-y')^2+u]}\right)}_{\text{depth-dependent kernel}} dx' dy' du \\
    \label{eq:depth-dependent-psf}
    &= \iiint_\Omega \frac{1}{2\sqrt{u}}\rho(x', y', \sqrt{u})\cdot f(v; \mu', \sigma') dx' dy' du,
\end{align}

\noindent where $\mu'=(x'-x)^2+(y'-y)^2+u$ and $\sigma'=\sigma c\sqrt{\mu'}$. When pulse width is non-zero, under the LCT transformations, the image formation becomes a depth-dependent convolution because $\sigma'$ is dependent on $\mu'$. As a result, the model is not strictly shift-invariant along the $z$ direction. Specifically, $\sigma'$ increases with depth, suggesting that our ability to localize far away objects decreases with depth. However, the peak locations of both the idealized kernel in \cref{eq:lct} and the depth-dependent kernel in \cref{eq:depth-dependent-psf} are the same. Therefore, we could still approximate the kernel as being depth-independent for smaller depths. We set the pulse width in LCT space $\sigma'$ by assuming a depth of $z=1$.  Empirically, we find this approximation to hold well for the range of depths we consider. However, for larger object motions in the $z$ direction, the effects of pulse width may become more important to model. 



\section{Analysis}

\subsection{Benefits of Known Object Shape as for NLOS Tracking}

The idea of using camera motion or object motion to improve reconstructed resolution is a well-known idea in synthetic aperture radar \cite{sherwin1962some} and inverse synthetic aperture radar \cite{ozdemir2021inverse}. However, in NLOS tracking, knowing the object shape is also a useful prior. A wide synthetic aperture provides spatial diversity in the measurements, which is important for NLOS reconstruction \cite{buttafava2015non, liu2019analysis}. However, if our goal is to track the position of an extended object, we can also leverage the spatial information temporally multiplexed in the measurements. 

Formally, we can understand this benefit by interpreting the tracking problem as a spatio-temporal de-convolution problem formulated by \cref{eq:tracking_as_convolution}. Here, our goal is to recover the shift $(\mathbf{\Delta}_x, \Delta_v)$ of the kernel, or the canonical measurement, $\mathcal{I}_0(\mathbf{x}, v)$. If the object shape is known, then $\mathcal{I}_0$ is known and the de-convolution is more well-posed. Tracking with an unknown shape is analogous to a blind de-convolution problem, where the PSF kernel is unknown. In such cases, either iterative blind de-convolution can be performed, or an approximate kernel can be used (e.g. the kernel corresponding to a point object). Using an approximate kernel would result in higher uncertainty in the recovered object position, whereas knowing the exact kernel would enable more fine-grained localization. 



\begin{figure}
    \centering
    \includegraphics[width=0.7\linewidth]{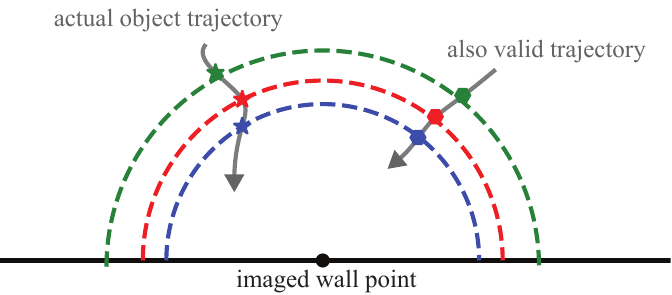}
    
    \caption{\textbf{Ill-Posed Tracking with Small Objects and Small Synthetic Aperture.} We use the keyhole imaging configuration as an example, where the synthetic aperture is just a single point, and the target hidden object is a point. Under this scenario, the object position cannot be well-constrained even if information is aggregated across multiple frames.}
    \label{fig:keyhole-ambiguity}
\end{figure}

\paragraph{Connection to Keyhole Imaging \cite{metzler2020keyhole}.} The idea of an extended object shape being useful for small synthetic apertures can be better understood in the context of keyhole imaging. Assuming our goal is to track a known object with only a single laser and pixel (i.e. the aperture is a single point), tracking a small object would be very challenging. To illustrate this idea, consider a point object moving along a trajectory. In the first frame, the camera captures a measurement $i_1=\mathbb{R}^{n_\tau}$ when the object is at $\mathbf{x}_1$. That measurement would constrain the object location to be along a hemisphere around the imaged point on the wall. Now, if the object were to move to a new position $\mathbf{x}_2$ in the next frame, the space of possible object locations is a hemisphere with same center but different radius. Any trajectory that places the object positions on these two hemispheres would be a valid trajectory, as shown in \cref{fig:keyhole-ambiguity}. The keyhole imaging work mitigates this issue by imaging objects with larger spatial extent, taking advantage of the spatial information encoded as a multiplexed temporal measurement. 


\subsection{Multi-Bounce Light as Opportunistic Blur}

Wbile the primary focus of this work is on non-line-of-sight imaging with consumer LiDARs, we also point out an advantage of using multi-bounce light even in line-of-sight scenarios. If a LiDAR was imaging a line-of-sight object with the goal of reconstructing it's 3D shape, the image formation is 

\begin{equation}
    i(\mathbf{x}, \tau) = \sum_{\mathbf{u} \in \mathcal{U}} \underbrace{\delta(\mathbf{u}-\mathbf{K}\mathbf{x})}_{\text{spatial sampling}} \cdot \underbrace{\rho(\mathbf{x}) \delta\left(\vert\mathbf{x}\vert-\frac{c\tau}{2}\right)}_{\text{scene response}},
\end{equation}

\noindent where $\mathcal{U}$ is the set of all pixels in the camera, $\mathbf{K}$ is the camera intrinsic matrix, and the 3D spatial coordinates $\mathbf{x}$ are in camera coordinates. From sampling theory, we know that the ability to recover the object shape $\rho(x)$ is determined by the spatial sampling rate, which is determined by the number of pixels $\vert\mathcal{U}\vert$. If the number of pixels are sparse (e.g. in low-power, resource-constrained settings), the spatial measurements of the object will be aliased. 

In such cases, one way to reduce the spatial aliasing is by using anti-aliasing filters in the optical domain, either by intentionally de-focusing the lens or increasing the sensor's instantaneous field of view \cite{duarte2008single, behari2025blurred}. Another way to intentionally introduce blur into the signal is by leveraging multi-bounce light that interacted with the object. We know from \cref{eq:cnlos_forward_model} that multi-bounce light results in a spatio-temporal blur of the object's shape with blur kernel $h$. As a result, sampling this multi-bounce signal (by using a nearby relay surfaces) can be beneficial from a reconstruction perspective depending on the object's spatial frequencies and the number of pixels available on the sensor.

\section{Implementation Details}

\subsection{Mechanical Gantry for Systematic Object and Camera Motion}

\paragraph{Quantitative Evaluation.}
For quantitative tracking and camera localization results in Fig.  \textcolor{blue}{1} of the main text, we use a mechanical gantry as a translation stage to move the object to fixed locations. For evaluation purposes, we use a stop motion capture at $10 \times 10$ locations along the trajectory of the object and camera. For results without quantitative evaluation, the data is captured in real-time. 

\paragraph{Reconstruction Results.}
We also use the gantry to move the camera to get a larger synthetic aperture to get full 3D shape of hidden objects. Specifically, we use the camera motion for the mannequin reconstruction in Fig. \textcolor{blue}{1} and the diffuse reconstruction in Fig. \textcolor{blue}{5} of the main text. In principle, we could use natural handheld motion instead of a mechanical gantry to reconstruct these objects if we had reliable camera pose.

\subsection{Experimental Setup}

\paragraph{Camera Setup.} We use a smartphone-grade LiDAR camera with $\sim 100$ pixels. Each pixel has a corresponding laser spot, and the laser and pixel illuminate and image along the same direction similar (i.e. confocal capture). All pixels and laser spots are active simultaneously. 

\paragraph{Use of Retroreflective Object.} For many of our experiments, we cover the hidden object in retroreflective cloth to preserve the confocal image formation model. In typical confocal imaging setups \cite{o2018confocal}, the laser/pixel pair are sequentially scanned to each point on the wall. As a result, light from one laser spot position will not interfere with the measurement at another detector position. However, such scanning is not possible with consumer LiDARs because they emit all laser spots simultaneously. However, by using a retroreflective material, light that originates at one laser spot can only interact with the hidden object and return directly back to that same spot. As a result, there will be no cross-mixing of light between different laser spots. This reflectance profile also enables capture of higher SNR signals. However, in Fig. \textcolor{blue}{5} of the main text, we also show that these sensors can image diffuse objects with our algorithm.

\paragraph{Camera Calibration.} For experiments where the camera is static, we place a checkerboard pattern on the wall and calibrate the camera pose with respect to the RGB camera. We treat the pose of the RGB camera and the LiDAR to be equivalent. 

\section{Supplementary Results}

\subsection{NLOS Tracking Results.} 

\begin{figure}
    \centering
    \includegraphics[width=0.7\linewidth]{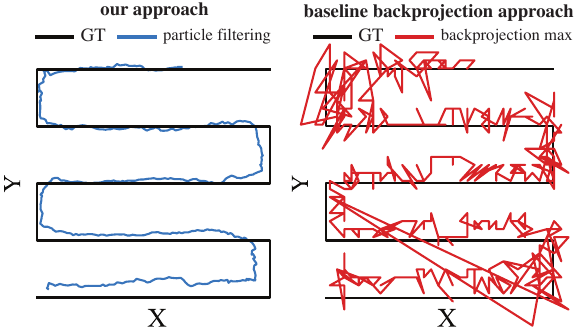}
    \caption{\textbf{Baseline Tracking Result.} We compare our tracking technique to a baseline technique of computing the filtered backprojection reconstruction of the hidden volume then computing the maximum value of the volume. We find that our particle filtering approach is more robust to noise and more computationally efficient.}
    \label{fig:supp_bp_baseline}
\end{figure}

\paragraph{Real-Time Diffuse Tracking.} The diffuse tracking result in Fig. \textcolor{blue}{5} of the main text is available as a video result. Each frame computes the kernel density estimation (KDE) of the particles to produce a probability distribution for the object position (plotted as a contour map). We also compare the tracking result in Fig. \textcolor{blue}{1} of the main text to a simple backprojection baseline in \cref{fig:supp_bp_baseline}. The baseline algorithm is to reconstruct the hidden scene, then compute the voxel location with maximum intensity. While backprojection doesn't outright fail, the tracking results are substantially noisier because they don't leverage any motion priors. Furthermore, they are computationally more expensive than our particle filtering approach because we use $1000$ particles whereas reconstruction requires computation over a 3D voxel cube ($30\times30\times 30$ in this baseline). These voxels also can only model discrete object positions, whereas particles can model continuous object positions. 

\begin{figure}
    \centering
    \includegraphics[width=0.7\linewidth]{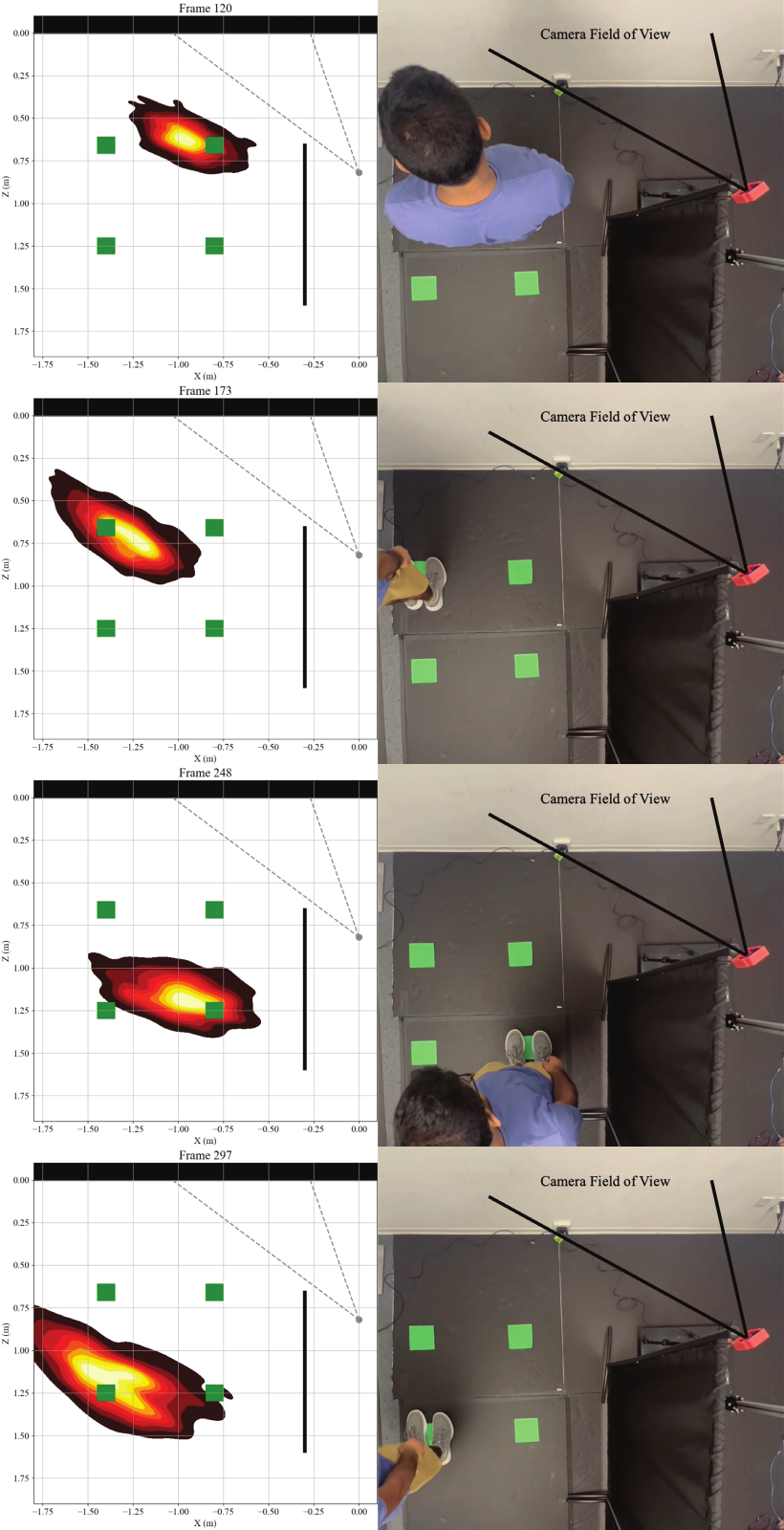}
    \caption{\textbf{Real-Time Tracking with ST VL53L8CX SPAD.} We demonstrate NLOS imaging with off-the-shelf hardware and minimal calibration. Available as video result.}
\label{fig:supp_st_tracking}
\end{figure}

In \cref{fig:supp_st_tracking}, we also demonstrate similar results with the ST VL53L8CX \cite{stspad}, an \textit{already commercially-available device}, and minimal calibration requirements. In particular, all we need to do for a data capture is (1) point the camera at a relay surface, (2) calibrate the point cloud of the planar surface, and (3) apply our particle filtering tracking algorithm. We believe that such demonstrations of plug-and-play NLOS imaging capabilities promises its feasibility in mainstream computer vision. On the reconstruction side, note that the uncertain regions are well-modeled implicitly by particle filtering. When the object is further from the aperture, the tracking resolution will degrade, as is well-known in NLOS imaging \cite{nam2021low}. However, particle filtering naturally allows for a mechanism to model these regions of ambiguity where multiple object positions are consistent with the measurements.   

\begin{figure}
    \centering
    \includegraphics[width=0.9\linewidth]{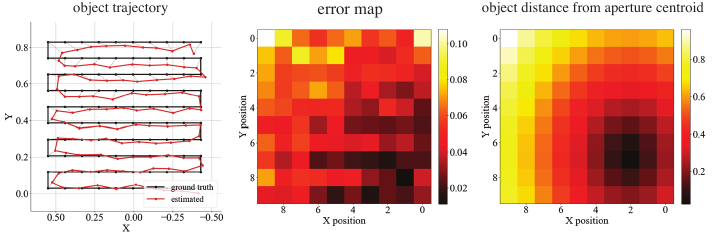}
    \caption{\textbf{Quantitative Tracking Results.} The $x-y$ trajectory of a hidden patch is shown. We observe that the tracking error is strongly correlated to the relative position of the object and virtual aperture. We validate this relationship by observing the similar structure between the error map and the aperture distance map, which computes the distance of the object to the aperture's centroid.}
    \label{fig:supp_quant_tracking}
\end{figure}

\paragraph{Quantitative Results.} We demonstrate quantitative tracking results by placing a $25 \times 25$ cm retroreflective patch on a mechanical translation stage. The translation stage moves along the $x-y$ plane parallel to the wall on a $10 \times 10$ grid. At each position, a measurement is captured of the NLOS object (in this case, a retroreflective patch). This setup emulates a stop motion capture of the moving scene. The tracking results are shown in \cref{fig:supp_quant_tracking}. The average position error over the entire trajectory is $4.7$ cm. However, we observe higher error in certain parts of the trajectory, as shown in the error map. This spatial-dependent error occurs due to the object's position with respect to the virtual aperture. As the object moves further from the imaged area on the wall, the tracking resolution worsens because the region of ambiguity increases. For this result, because we captured stop motion, we could capture fewer frames. As a result, we had to loosen the constraint on our motion model to allow for a larger step size between subsequent frames (we use $r=20$ cm). However, the stop motion capture was also helpful because it enabled. This result could be interpreted in terms of the recovered tracking resolution, ignoring SNR considerations.

\subsection{Camera Localization Results}

\begin{figure}
    \centering
    \includegraphics[width=0.7\linewidth]{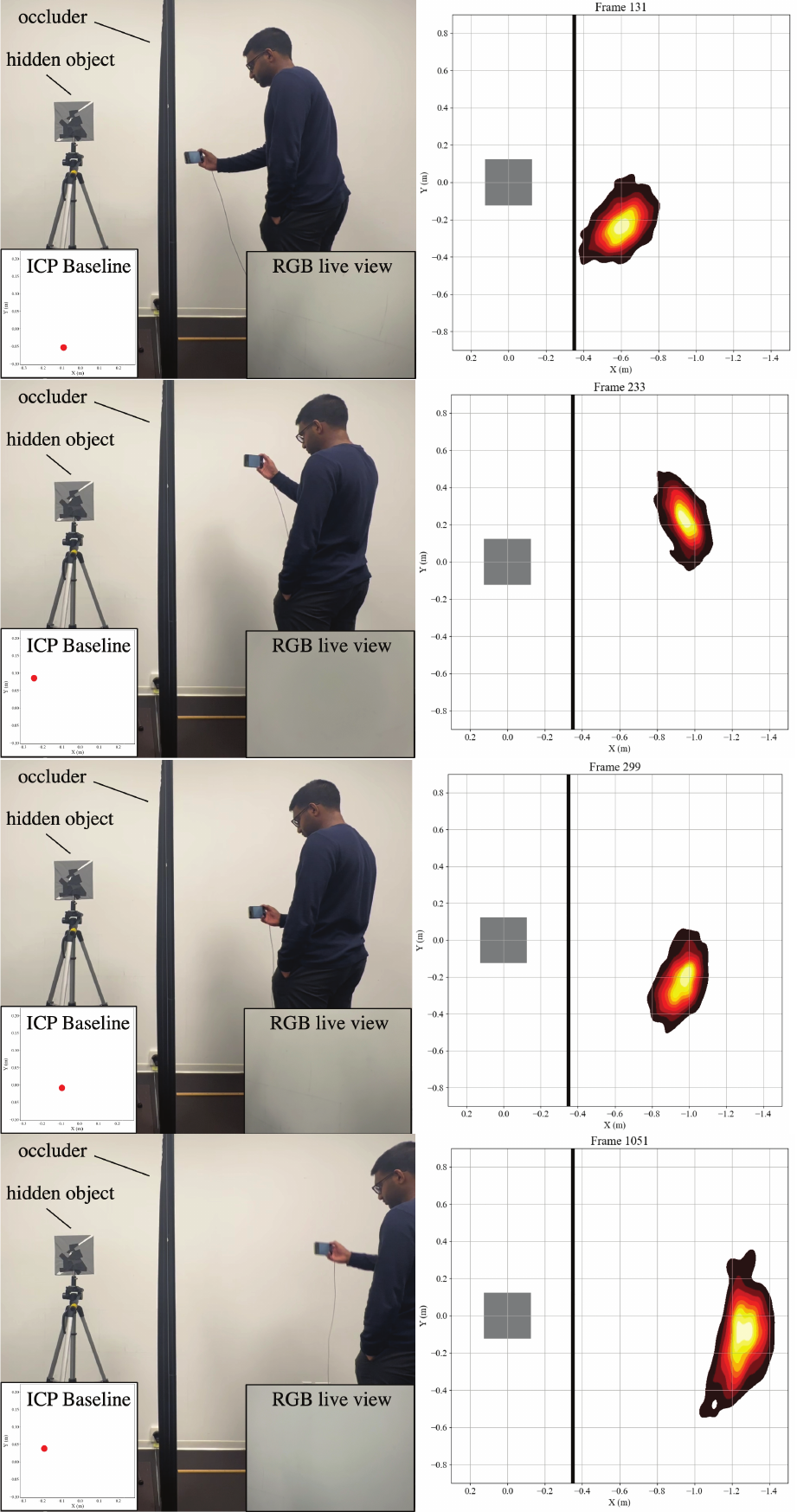}
    \caption{\textbf{Real-Time Handheld Camera Localization.} The camera has unstructured 5D motion and is observing a textureless wall, which would be challenging for conventional odometry techniques such as ICP. However, using the hidden object as a visual cue, we are able to recover the camera's trajectory. Available as video result.}
    \label{fig:supp_cam_local_rt}
\end{figure}

\paragraph{Real-Time Handheld Capture.} We perform a real-time demonstration of camera localization under unstructured handheld camera motion. This result is also available as a video. xThroughout the capture, the camera is imaging a planar white surface that contains little-to-no visual features. We show that we are able to recover the camera's trajectory in \cref{fig:supp_cam_local_rt}. Conventional techniques such as iterative closest point (ICP) due to limited line-of-sight geometric and texture variation. In our capture setup, we assume only 5D motion, where we assume the camera does not roll parallel to the wall. In principle, our method could support roll rotation and obtain the 6D camera pose by modeling the camera roll as part of the state space (in addition to the $(x, y)$ translation. The  only restriction is that the hidden object shape must not be symmetric in the $x-y$ plane. Our use of particle filtering also enables robust and efficient modeling of the uncertainty distribution of the camera position. In the bottom row of \cref{fig:supp_cam_local_rt}, the camera is far away from the hidden object. As a result, the resolution of the recovered camera location will be much lower due to the distance of the object from the synthetic aperture. The distribution of the particles reflects this ambiguity by increasing it's region of uncertainty. Specifically, ambiguity here refers to cases where many different camera positions are consistent with the measurement. When the camera's field of view is closer to the hidden object, the region of ambiguity decreases because fewer object positions are consistent with the measurement. A similar observation can be made for the object tracking case as well. 


\bibliographystyle{ieeetr}
\bibliography{bib}

\origbibliographystyle{ieeetr}
\origbibliography{bib}

\end{document}